# Self-harm: detection and support on Twitter

Muhammad Abubakar Alhassan[1], Isa Inuwa-Dutse[2], Bello Shehu Bello[3] and Diane Pennington[1]
[1]Department of Computer and Information Sciences, University of Strathclyde, Glasgow UK
[2]Department of Computer Science, Federal University Dutse, Nigeria
[3]Department of Computer Science, Bayero University Kano, Nigeria

**Abstract:** Since the advent of online social media platforms such as Twitter and Facebook, useful health-related studies have been conducted using the information posted by online participants. Personal health-related issues such as mental health, self-harm and depression have been studied because users often share their stories on such platforms. Online users resort to sharing because the empathy and support from online communities are crucial in helping the affected individuals. A preliminary analysis shows how contents related to non-suicidal self-injury (NSSI) proliferate on Twitter. Thus, we use Twitter to collect relevant data, analyse, and proffer ways of supporting users prone to NSSI behaviour. Our approach utilises a custom crawler to retrieve relevant tweets from self-reporting users and relevant organisations interested in combating self-harm. Through textual analysis, we identify six major categories of self-harming users consisting of inflicted, anti-self-harm, support seekers, recovered, pro-self-harm and at risk. The inflicted category dominates the collection. From an engagement perspective, we show how online users respond to the information posted by self-harm support organisations on Twitter. By noting the most engaged organisations, we apply a useful technique to uncover the organisations' strategy. The online participants show a strong inclination towards online posts associated with mental health related attributes. Our study is based on the premise that social media can be used as a tool to support proactive measures to ease the negative impact of self-harm. Consequently, we proffer ways to prevent potential users from engaging in self-harm and support affected users through a set of recommendations. To support further research, the dataset will be made available for interested researchers.

## 1. Introduction

Self-harm is an umbrella term that could denote any act or intentional inimical behaviour to oneself. Because inimical behaviour could be subjective, there are various meanings associated with self-harm (self-injury) within the research community. Another contributing factor to the diverse concepts of self-harm is because researchers use different terminologies to convey the meaning of the behaviour without considering the direct or indirect nature of it. Self-injury is described as any behaviour performed deliberately that could lead to psychological or physical harm to oneself with no intention to die . In other words, non-suicidal self-injury (NSSI) is the direct, intentional damage to one's own body tissue with no intention to die (Nock, 2010). According to Nock (2020), a direct self-harm can be from actions such as cutting or burning of the skin while taking an excessive drug overdose is a form of indirect self-harm leading to poor health due to chemical changes in the body. The work of (Madge et al. 2008) defined self-harm as a non-fatal act in which a person intentionally engages in a behaviour such as cutting, jumping from an extended height distance, illicit drugs ingestion, taking of non-ingestible substances, or overdose of legal substances with the intent of causing harm.

Regardless of the intention, self-harming is, among other reasons, considered by some as a means to cope with unbearable feelings. Thus, it is pertinent to ask what is the motivation for someone to engage in *self-harm?* Researchers have studied some of the reasons that cause people to consider self-harming to cope with difficult emotions or situations. A study by (Madge et al. 2008) outlined eight such possible reasons based on the responses from young people across seven European countries. Although the participants in (Madge et al. 2008) have the option to choose multiple reasons, it was not common for a single reason to be attributed to an episode of self-harm. However, the following reason "*I wanted to get relief from a terrible state of mind*" was reported by about 71% of the participants. This observation was also corroborated in the work of (Rasmussen et al., 2010). Moreover, interpersonal reasons such as "*I wanted to frighten someone*" and "*I wanted to get some attention*" have been identified as the reasons for engaging in self-harm behaviour (Rasmussen et al., 2010). The responses given by the participants in both (Madge et al. 2008) and (Rasmussen et al., 2010) studies have been attributed to a psychological state of the self-harming individual. We surmise that if the motivation is to self-harm for the sake of attention, which is at the detriment of one's well-being (Madge et al., 2008), then the modern-day social media offers an avenue for users to self-harm and post the details.

**Online social media.** As the name implies, social media is an online tool that promotes social interactions among users through creating and sharing multimedia and non-multimedia contents. There is no gainsaying in stating that the advent of online social media has revolutionised various aspects of our social lives owing to the popularity of platforms such as Twitter, Facebook, Snapchat, TikTok, Baidu, and WhatsApp. The online social media platforms offer useful resources to study various social phenomena, such as self-harm detection and support. For instance, the National Health Services (NHS) in the United Kingdom listed some useful organisations that support affected individuals, mostly via online social media platforms. Among these platforms, Twitter is one of the leading social networking sites that allow instant online social support from organisations offering mental health services such as YoungMinds[1] and Samaritans[2]. These organisations aim to ease anxiety among disturbed individuals and disseminate useful support information for people suffering from mental health issues such as self-harm. Twitter is one of the channels through which they reach out to the broader communities and offer support. The instant nature of online interaction makes it a suitable option to seek support. Distressed individuals seeking online help were found to be more at the risk of self-harming when compared to those seeking offline help from medical professionals (Mitchell and Ybarra, 2007). The impact of sharing NSSI-related content can be viewed as a double edge sword: (1) it enables the study of self-harm through data analysis and (2) exposes other users to the influence of such content. Motivated by the former, our goal in this study is to examine NSSI-related contents on Twitter and answer the following research questions.

1.1  *Research Questions*
To achieve the study's goal, we put forward the following questions for investigation:
- *What is the nature of the discussion about self-harm on Twitter*? Here, we aim to understand how discussions about self-harm are unfolding, and to study the discussants. In line with previous work (De Choudhury et al., 2013), we are interested in examining behavioural cues with respect to the users' posts, focusing on analysing the expressed emotion or sentiment and linguistic styles.
- *What are the strategies used by relevant organisations to support and promote self-harm recovery on Twitter*? Using longitudinal data from the Twitter accounts of self-harm support organisations, our goal is to reveal and analyse various strategies used by the organisations and identify common best practices.
- *How followers of the applicable support handles react to the information they received overtime?* Intuitively, this question focuses on finding the relationship between the support handles and online users seeking support against self-harm. This will also make it possible to study the frequency of social activity of relevant accounts.

The remaining part of the paper is structured as follows. Section 2 presents some of the existing relevant studies, Section 3 offers a detailed description of the data collection and some preliminary analyses. Section 4 reports answers to the research questions and their implications. Finally, Section 5 summarises our main findings and suggests some key areas for future research.

## 2.  Related work

Depression is one of the common mental health issues that is associated with self-harm, and a study found that 1 in 6 people may have experienced depression in the past week (McManus et al., 2016). In this age of hyperconnectivity, online social media platforms such as Twitter are useful in measuring and detecting depressed users (De Choudhury et al., 2013). Past studies reported that a significant number of self-injurers utilise the internet and are very likely to involve in online behaviours that could put them at risk (Mitchell and Ybarra, 2007). We review relevant studies on self-harm and online social media.

---

[1] https://youngminds.org.uk/
[2] https://www.samaritans.org/

2.1  *Self-harm and Social Media*

Nowadays, the number of social media users is increasing rapidly, resulting in a large amount of unstructured data worthy of investigating. In the context of this work, the unstructured data is being utilised for various behavioural analyses and changes in mental health condition. For instance, the work of (De Choudhury et al., 2014), reported how social media users seek and share health-related information with others, and (Bazarova et al. 2015) studied how online users anonymously express their emotions and experiences. While many studies have examined the impact of social media concerning public health challenges, however, for people with mental health issues such as self-harm, platforms such as Twitter facilitate the formation of communities of individuals with similar problems or interests, exchange of peer support and sharing of experiences (Hilton, 2017). Although Twitter is considered to be useful for seeking online help by users engaging in self-harming (Hilton, 2017), some existing studies believe that the social media environment is not safe for this community. This is due to the danger of being exposed to other ways of self-harming and triggering contents that could normalise self-harm behaviour (Brown et al., 2018).

Contents related to self-harm in online social media platforms, notably Twitter, Tumblr and Instagram, can be broadly categorised as follows: graphic content, negative self-evaluations, references to mental health, discouragement of deliberate self-injury, and recovery-oriented resources (Miguel et al., 2017). Similarly, the work of (Xian et al. 2019) utilises NSSI-related datasets from Instagram to identify and classify images as NSSI or non-NSSI. . Noting the importance of identifying self-harming content and subsequent prevention, it will be crucial to complement the existing studies with textual data from Twitter. Additional effort is needed to investigate how digital networks facilitate the self-harm recovery process (Hilton, 2017). To help in curtailing online self-harming attitude, especially noting how depressed users are more likely to be inclined towards self-attention and increasing negative emotions (De Choudhury et al., 2013), we offer a comprehensive analysis of self-harm datasets from Twitter.

## 3. Methods

In line with the previous approach (De Choudhury et al., 2013), we are interested in examining behavioural cues using the users' posts to answer the questions listed above (Introduction section). Thus our approach is centred around the following: (1) identify a suitable online social media platform for data collection (2) how to identify NSSI-related content, and (3) preliminary analysis involving categorisation of NSSI users, trend analysis and visualisations.

3.1  *Data Collection*

The advancement in digital social networks has provided the opportunity to access enormous datasets for various reasons. Social networking sites gain the highest number of users worldwide in which they provide their information, thoughts and opinions through various channels, such as mobile phones, laptops, and tablets (Kumar et al., 2014). The data generated from social networks like Twitter offers essential mixture and diverse formats of different sizes that makes it flexible and suitable for researchers to conduct their investigations (Steinert-Threlkeld, 2018). In other words, Twitter has always been an excellent data source for many academic researchers from different fields. The diverse data accessible via the platform's application programming interface (API) offers a comprehensive overview of the user's behaviours or actions and the contents they produced. Motivated by the limited number of studies exploring the sources of online social help for NSSI-related issues, this study chose Twitter for data collection. We created a Twitter developer account and used secured access keys and tokens in obtaining the data. We created a crawler that collects tweets using defined keywords, hashtags, and tweets directly from Twitter accounts without violating any of the terms and conditions of the Twitter's API. The data collection process is given in phases.

3.1.1  *Phase One*

In the first phase of the data collection, we manually searched Twitter and identified the list of organisations that provide support for self-injurers and people with mental health problems. We retrieved a list of relevant organisations from the official website of the National Health Service (NHS)[3] in the United Kingdom as of 21 december 2019. Moreover, we also utilise other sources of mental health support on Twitter using a snowballing technique. However, because our study is focused on Twitter platform, we limited our investigations to organisations that are using a Twitter account in reaching out to the general public as listed in Table1.

---

[3] https://www.nhs.uk/conditions/self-harm/

**TABLE 1** List of Support handles

| Organisation name | Twitter support handle | Followers count |
|---|---|---|
| Mind | @MindCharity | 461,665 |
| Selfharm UK | @selfharmUK | 4,409 |
| Shout UK | @GiveUsAShout | 9,476 |
| Self injury Support | @sisupportorguk | 3,216 |
| LifeSIGNS & | @LifeSIGNS | 3,412 |
| Mental Health Notes | @depressionnote | 192,644 |
| The WISH Centre & | @TheWISHCentre | 3,946 |
| Samaritans* | @samaritans | 132,939 |
| Stop Self Harm | @stopselfharm | 21,290 |
| YoungMinds* | @YoungMindsUK | 164,966 |

\* Organisations listed by the NHS

#### 3.1.2  Phase Two

In the second phase of the data collection process, we utilised *#selfharm* hashtag to retrieve relevant tweets from Twitter. Although searching for online self-harm contents can be done using a keyword like *self-injury* or *self-harm*, the *self-harm hashtag* appears to be the most widely used in eliciting contents related to self-harm on social media. Through the proposed approach, we obtained a set of *8,063* tweets from unique users with *7,914* replies from 15/04/2020 to 19/05/2020. We separated the languages used by users in tweeting about self-harm and discovered that tweets written in the English language represent a high proportion of the tweet, accounting for about 98% of the tweets. Therefore, we conducted our analysis on tweets that are explicitly written in English language.

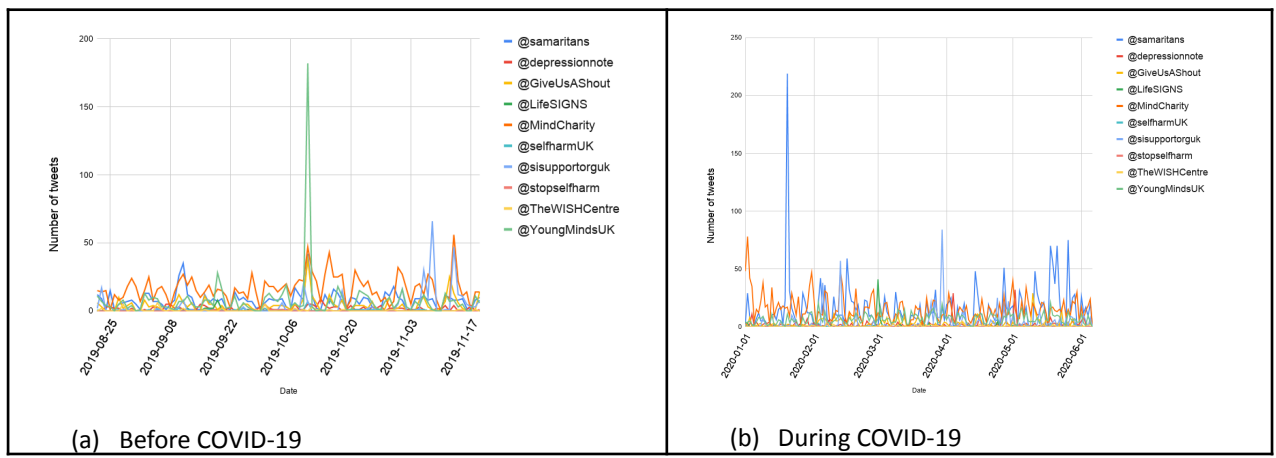

(a) Before COVID-19  (b) During COVID-19

**Figure 1:** A summary of the tweeting activity of the *self-harm* support accounts or handles on Twitter. The visualisation is given as pre-COVID-19 and during the COVID-19 outbreak. Although spikes in the activities are roughly the same, the activity during COVID-19 is comparably higher

### 3.2  Preliminary Analysis

#### 3.2.1  Basic statistics

Based on 7-days interval, Figure 1 presents information about the tweets' distribution over the collection periods. Although the first and last week of the collection period showed that the number of tweets received per day is less than 200, we found that users post, on average, 237 tweets per day. Both graphs in Figure 1 show spikes that revealed meaningful information. In Figure 1 (a), we found that the number of tweets from the *@YoungMindsUK* account exceeds two hundred on 10/10/2019; this date corresponds to global mental health awareness day. This indicates that the *@YoungMindsUK* account is socially active in reaching out to users and raising awareness about mental health on Twitter.

Similarly, Figure 1 (b) shows that many tweets were received from the *@samaritans* account on 20/01/2020, which is the day of the blue Monday campaign of the year. Although blue Monday day is chosen as the third Monday of January every year, it is described as the most depressing day (Cheater, 2019). However, Our analysis found that the *@samaritans* account became very engaged in reaching out to many users and campaigning about blue Monday in 2020. In a nutshell, the most active handles between August and November 2019 (before COVID-19 in the UK) were *@MindCharity, @Samaritans,* and *@YoungMindsUK*. Similarly, we discovered a swap between these accounts in the first half of the year 2020 as the *@Samaritans* overtook the *@MindCharity,* then the *@YoungMindsUK* profile; these support handles often use hashtags to post information on Twitter. Also, users can use the same hashtags to explore contents associated with the hashtags.

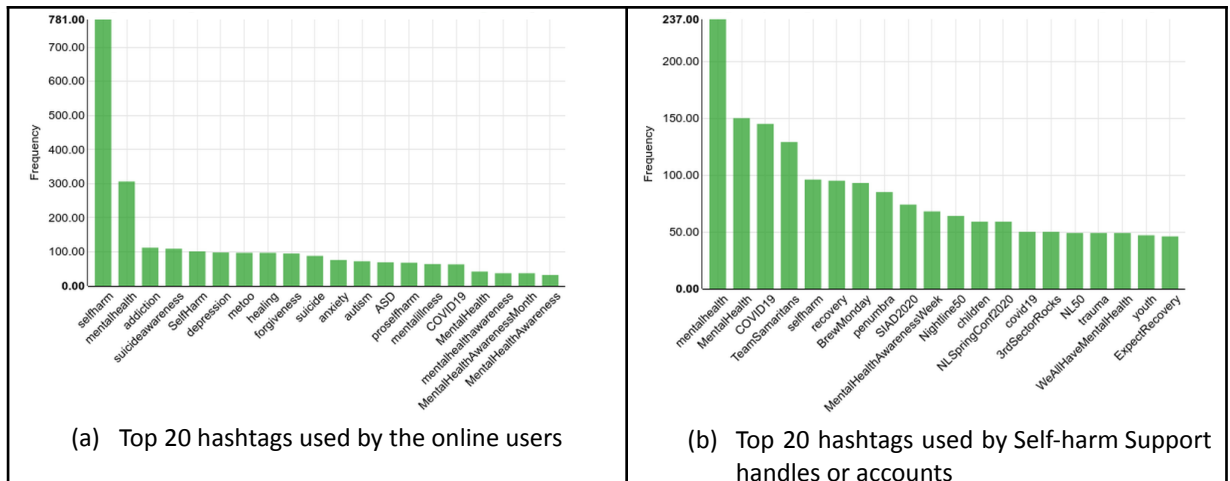

(a) Top 20 hashtags used by the online users

(b) Top 20 hashtags used by Self-harm Support handles or accounts

**Figure 2:** A visualisation of the top hashtags used by various accounts concerned with *self-harm* on Twitter

Although our study retrieved tweets directly from the support handles, we analysed the top 20 hashtags that appeared on the handles' tweets. Figure 2 shows the top 20 hashtags used by the support handles in interacting with online users. We can observe that neither *#selfharm* nor *#selfinjury* makes it to the top 20 hashtags. The most common hashtags were *#BrewMonday* and *#mentalhealth*. Even though *#RedJanuary* hashtag appeared in most of the tweets posted by the handles, it is clear that these handles were actively raising awareness and information using mental health hashtags and other hashtags related to the global health crisis (COVID-19). Furthermore, we used the tweets we received through *#selfharm* hashtag to understand the common hashtags used by the online users in tweeting or information about self-harm. In addition to these hashtags, our analysis found that users use specific hashtags related to *depression, anxiety*, and *autism* and all these are associated with self-harm. Also, users frequently use the *#addiction* and *#suicide* hashtags to tweet about self-harm behaviour. Overall, we found that both support handles and users use similar hashtags in discussing issues related to mental health and well-being on Twitter.

### 3.2.2 User engagement
There are many ways to examine how users engage with each other on Twitter (Inuwa-Dutse et al., 2018). This study defines user engagement as any of the following actions: such as a tweet, retweet, and like used by a user to communicate or interact with other online users. Our experiment used essential metrics to understand how support handles are engaging with users on Twitter. These metrics measure online activities such as the *frequency of tweets* and *retweets*, *likes*, and *average sentiments* expressed by the support handles. As suggested by (Gilbert, 2014), we applied the valence aware sentiment reasoner (VADER), a rule-based approach for analysing textual information to understand the polarity (positive or negative) of a tweet from the support handles.

### 3.2.3 Topic Analysis
It is vital to understand the kind of topics users discuss on Twitter that are related to self-harm. To uncover such topics, we use the data we collected using the *#selfharm* hashtag. A applied a rudimentary preprocessing step to clean the data to ensure it does not contain irrelevant or function terms such as stopwords. Moreover, to better understand the data and inform the discussion, we utilised the Latent Dirichlet Allocation (LDA) algorithm (Blei et al. (2003) for the topic analysis. Noting the size of tweets, we focus on the top trigrams to understand the discussions' meaning and context (see Table 2 for the trigrams and categorisation of the users according to the learned topics).

**Table 2:** Some trigrams and corresponding themes from relevant tweets. We categorised the users (according to their tweets) based on the degree of similarity of their tweets to the topics learned via the LDA.

| User's category | Trigram | Example tweet |
|---|---|---|
| Anti-selfharm | encouraging self harm, suicide self harm, people self harm, mentalhealth suicide awareness, suicide awareness selfharm, traumatic experience, self harm addiction | 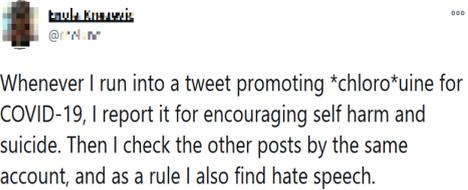 Whenever I run into a tweet promoting *chloro*uine for COVID-19, I report it for encouraging self harm and suicide. Then I check the other posts by the same account, and as a rule I also find hate speech. |
| Pro-selfharm | people self harm | 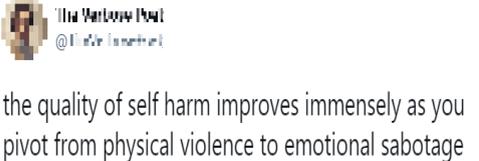 the quality of self harm improves immensely as you pivot from physical violence to emotional sabotage |
| Support seekers | selfharm addiction metoo, selfharm healing metoo, metoo forgiveness healing, autism selfharm asd | 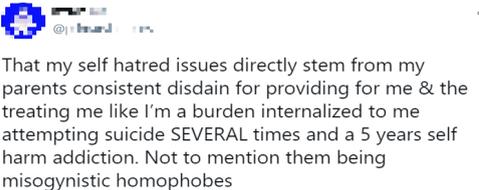 That my self hatred issues directly stem from my parents consistent disdain for providing for me & the treating me like I'm a burden internalized to me attempting suicide SEVERAL times and a 5 years self harm addiction. Not to mention them being misogynistic homophobes |
| Inflicted | tw self harm, self harm scars, self harm suicide, self harm im, self inflicted injury, used self harm | 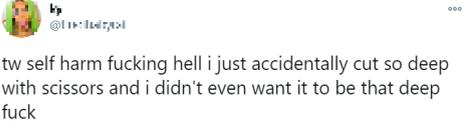 tw self harm fucking hell i just accidentally cut so deep with scissors and i didn't even want it to be that deep fuck |
| At-risk | like self harm | 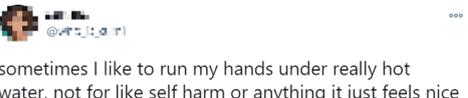 sometimes I like to run my hands under really hot water, not for like self harm or anything it just feels nice |
| Recovered | clean self harm, self harm free | 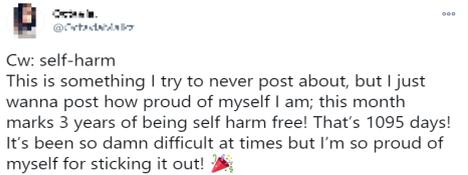 Cw: self-harm This is something I try to never post about, but I just wanna post how proud of myself I am; this month marks 3 years of being self harm free! That's 1095 days! It's been so damn difficult at times but I'm so proud of myself for sticking it out! 🎉 |

## 4. Result and Discussion

Twitter draws the attention of diverse users from different parts of the world to discuss self-harm and related incidences. Investigating the nature of the discussion will uncover useful insights that could be essential in safeguarding the online social space and in decision making. In this section, we present and discuss additional findings from the study.

### 4.1 Nature of the discussion about self-harm on Twitter

In analysing self-harm associated conversations on Twitter, we examined the set of tweets discussed in Section 3.1.2 and identified the key topics of the discussions. We grouped users based on similar topics, and our experiment found various discussion themes that are specifically related to self-harm. Based on the observed themes, we present the following six fundamental categories (*Inflicted, anti-self-harm, pro-self-harm, recovered, at-risk*, and *support seekers*) of users who participated in self-harm discussions on Twitter (see Table 2). For brevity, we extracted the top trigrams from the set of tweets to learn the meaning of the topics and compare them with the user's tweets. Users under the the *anti-selfharm* group tend to report any account they found encouraging other users to harm themselves or commit suicide;they advise users on how they can stop the behaviour and educate them about sources of help available.

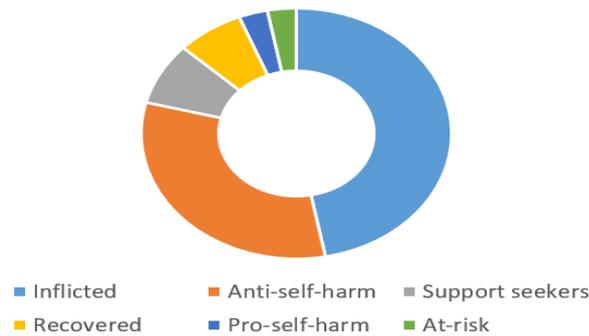

**Figure 3:** Categories of users discussing self-harm on Twitter

On the other hand, people seeking help expressed concern over how they have become addicted to self-harm and its healing process. Intuitively, there were fewer tweets from both *pro-selfharm* and *at-risk* users. This suggests that only a small proportion of these groups disclosed their experiences on Twitter social networks, and they exhibit a significant gap with other groups that are providing support. Some of the users in the *inflicted category* often use a *trigger warning* (tw) messages before posting information about their self-harm experiences, and the kind of injuries they suffered that led to scars. Users recovering from self-harm were tweeting to explain how they have become free and clean from self-harm. Consequently, our investigation found that, out of all users, the *inflicted* category has the highest percentage, as shown in Figure 3. This group of users reached up to 47%, indicating that nearly half of the investigated tweeters engage in discussing their self-harm experiences and its consequences such as scars. The *anti-self-harm* category of users that communicate about raising awareness on the issues and effects of self-harm, as well as seeking advice and support from health professionals have received 32%. This percentage reduced significantly to 3% for both *pro-self-harm* and *at-risk* groups. Furthermore, there is a simultaneous allocation rate between the group of users seeking support and those that recovered from self-harm behaviour.

4.2  *Support handles and online users*

Understanding how the online users react to the support handles' information is crucial to the analysis of the impact of the support handles on the online users. In this section, we study the online users' opinion on the support handle for six months.

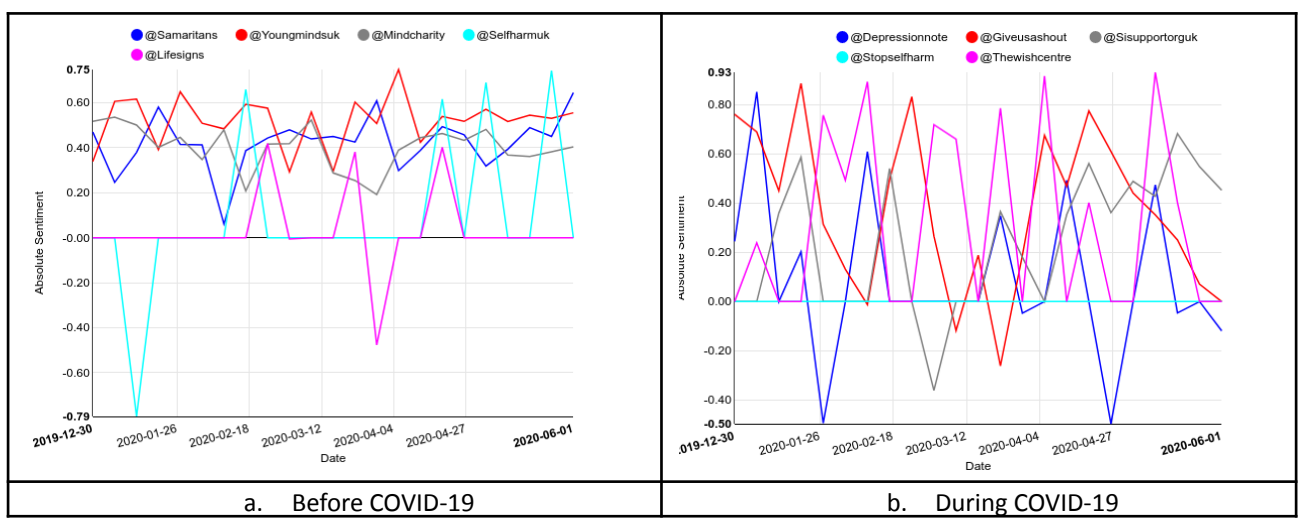

**Figure 4:** The dynamism in opinions of users on self-harm over time

Figure 4 shows the dynamism in opinions of users on self-harm-related topics on Twitter. To enhance readability, we split the opinion into two based on the variability of the absolute sentiment. The absolute sentiment shows the overall opinion of online users at a given time. We can observe some instances where there is a decline in the sentiment graph. Overall, the negative opinions are relatively higher than the positive opinions. They incline if positive opinions are higher than negative opinions.

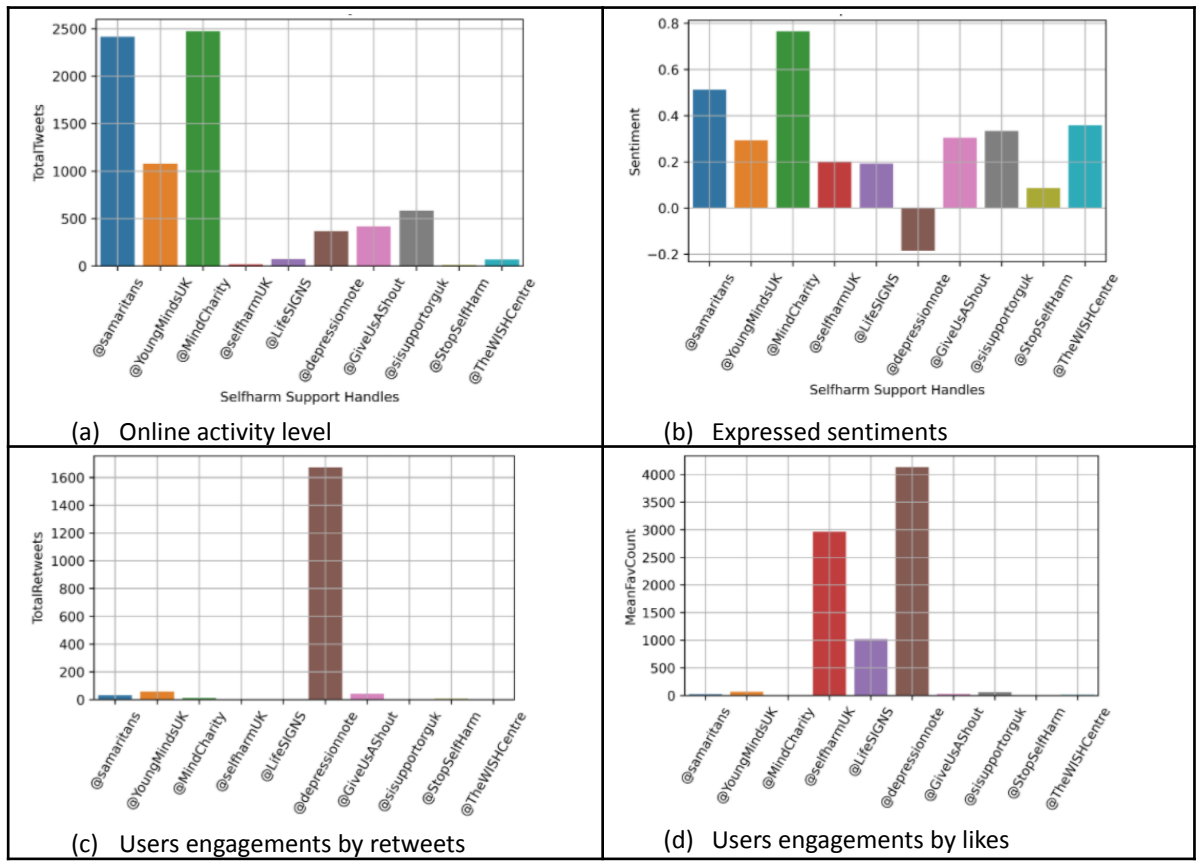

**Figure 5:** A visualisation of some of the engagement features that have been computed as a function of the number of *tweets, retweets count, favourites count or like, and sentiment analysis*

In Figure 5, sub-figure (a) measures how active the support handles are on Twitter; the activity level is calculated as the number of tweets posted during the collection period. Because we have each account's activity level, it is crucial to measure how the users respond to the posted contents by computing sentiments and engagement activity. In the expressed sentiments sub-figure (b), most of the accounts exhibit positive themes except *@depressionnote* handle. The negative value in the handle can be explained by the perceived negative connotations associated with some of the terms in the tweets. For instance, *feeling depressed?* is a question that would typically be followed by some useful tips afterwards. Similarly, sub-figures (c-d) report the engagement intensity according to information sharing (*retweets*) and *favourites* or *likes* associated with each handle's content.

4.3   *Self-harm Support Strategies on Twitter*
In this section, we applied the approach proposed by (Bello et al. (2018)) to understand the behaviour of Twitter accounts and the corresponding strategies to engage users, especially by the support handles Figure 5 depicts the tweeting strategy of *@Samaritans, @MindCharity* and *@YoungMindsUK*. Using the strategy, we can ask the following: *why @depressionnote is more engaging than the other handles as shown in Figure 5(d)*?. While the result includes basic constructs of an account's tweeting pattern, we will focus on the surface information Intuitively, the former (constructs) is more relevant to the study of automated accounts' automation pattern. Family of green colour in the result represents positive sentiment while that of red represents negative sentiment.

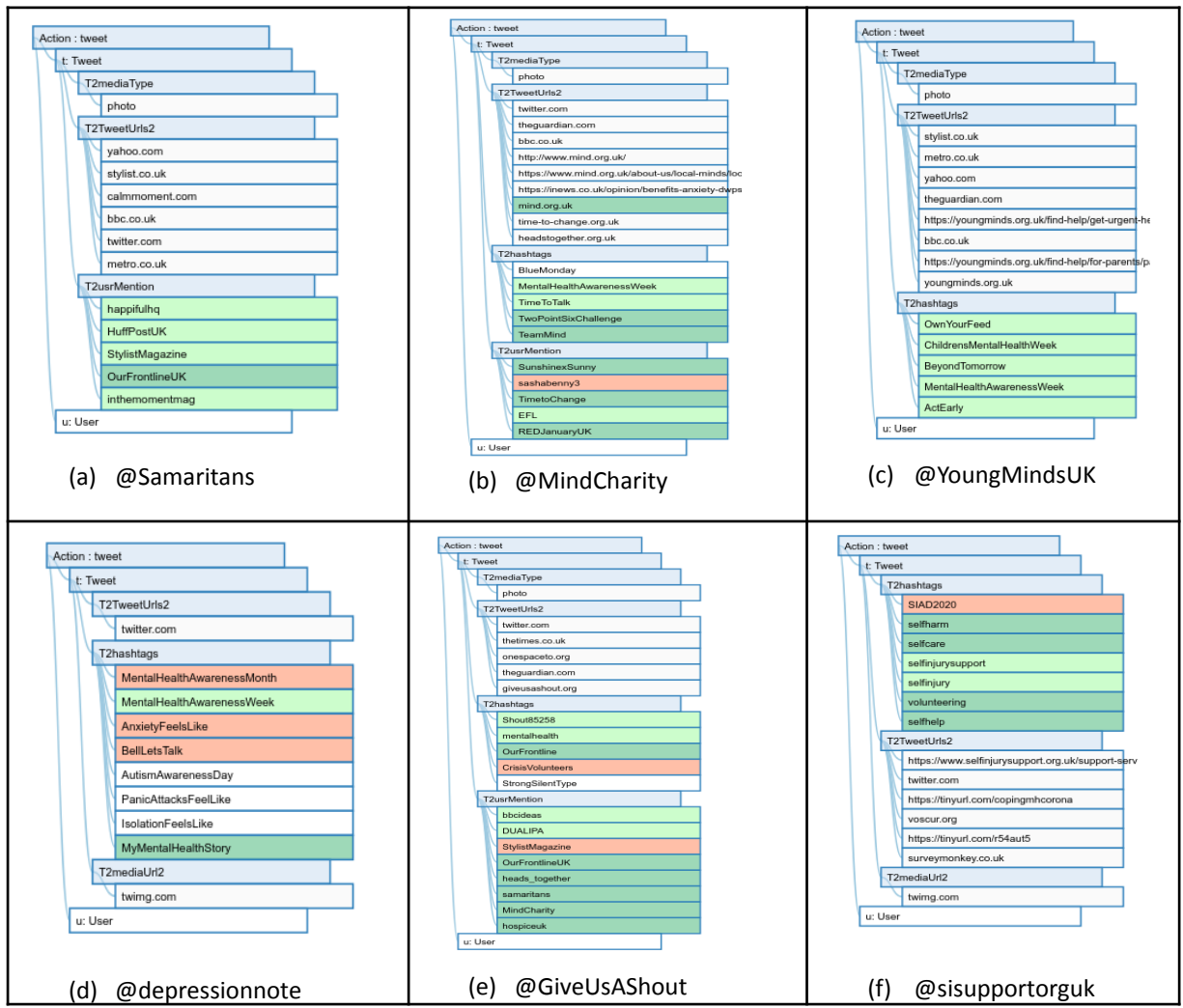

**Figure 6:** Tweeting strategy of the support handles

As shown in Figure 6, the *@Samaritans* account mainly posts tweets with photo and web-links from the news media groups such as *bbc.co.uk*, *metro.co.uk* and connects users to other support handles (*@happifulhq* and *@OurFrontlineUK*). Unlike the *@MindCharity* and *@YoungMindsUK accounts*, the *@Samaritans* is not using hashtags as part of its campaign strategy in reaching out to members of the online community. Although all the three most active accounts exhibit a typical pattern in terms of URLs sharings, only *@YoungMindsUK* and *@MindCharity* have a common hashtag (*#MentalHealthAwarenessWeek*) in sharing information and increasing awareness about mental health. In contrast to other support handles, *@MindCharity* combines both the styles of *@YoungMindsUK* and *@Samaritans*. This could be one reason the handle received more *likes* and increased user engagements. Furthermore, *@depressionnote* uses hashtags attributing to feelings of anxiety. Also, there is a change in emotions regarding the account's campaign using *#MentalHealthAwarenessWeek* and *#MentalHealthAwarenessMonth*. The former hashtag attracts positive sentiments while the later is associated with negative feelings. The use of mental health attribute tags such as *#AnxietyFeelLike, #IsolationFeelLike* could be the reason why *@depressionnote* attracts users engagement. The tactics used by *@GiveUsAShout* are different from the rest of the support handles because it connects users to two most active support handles (*@samaritans* and *@MindCharity*) on Twitter. Notwithstanding, the *@sisupportorguk* is the only support handle that used a unique hashtag, *#SIAD2020*, which stands for self-injury injury awareness day to share self-harm information with online members.

4.4 *Implication of the findings*
Researchers highlighted the influence of social networking sites on people who are harming themselves. They suggest that it is crucial for both mental health professionals and caregivers to evaluate self-injured people's online social behaviour (Lewis et al., 2012; Lavis and Winter, 2020). Applicable guidelines, especially in clinical settings, were presented to assess the nature and extent of online activity concerning self-injury (Lewis et al., 2012). Meanwhile, the extent and quality of support for those who access social platforms may directly affect their likelihood of accessing medical support. It is against this backdrop we investigated a set of online support for this group of people on Twitter. Our findings suggest that the Twitter online community members perceived the support, advice, and awareness from the support handles as helpful (positive). Our research contribution appends to existing studies on the impact of social networks on users that are self-harming. Mental health experts and academic professionals can benefit from the study by understanding the potential impact of self-harm discussions on Twitter. Moreover, open and informed conversations concerning self-harm, its representation on social networks, and the possible risks are crucial. Through this study, we uncover useful insights about the nature of self-harm discussions on Twitter, which could help medical practitioners understand and strategies for effective care. For instance, they could leverage the support handles to reach out to people who are harming themselves through Twitter and offer support effectively.

From an engagement point of view, the support handle's behaviour revealed essential areas of improvement. In contrast to less active handles, the most enthusiastic support handles are engaging in raising awareness about self-harm. User's opinion on information they received from the support handles indicates valuable feedback. For instance, the *Mental Health Notes organisation* (with the handle *@depressionnote*) needs to reconsider its campaign strategy for better engagement with the online users. Additionally, support handles could benefit from this study by improving their campaign strategy through sourcing information from medical institutions and educating people about the dangers of self-harm. This investigation confirms the need for clinicians to work hand in hand with support handles in preventing self-injury and other mental health problems.

## 5. Conclusion and future work

Self-harm is a severe mental health challenge that is growing in young people around the world. Although Twitter often offers helpful support for those who are harming themselves, the sense of societal identity, connection and empathy created by such support may normalise or reinforce the behaviour. Our study demonstrates the positive impact of social media for self-injurers seeking help on Twitter. Firstly, we approached the first research question through a mixed-method technique and discovered the key themes surrounding self-harm discussions on Twitter. We categorised different users participating in the NSSI talks by classifying them based on similar topics. Secondly, we employed reverse engineering methods to uncover the strategies used by self-harm support accounts on Twitter in offering social support to affected users. Thirdly, we analysed the online users' opinions towards support handles to understand their degree of satisfaction with the available support. The overall sentiment is positive, however, more is needed to curtail the menace of self-harm. We make the dataset available for further research.

The study's limitations are recognised, most notably because it is unclear whether the people who participated in the discussion were primarily from clinical or non-clinical cohorts. Although other support handles could be available, the study is limited to only ten support handles. Additional study may consider other sources of social support for self-injurers on social media. Even though young individuals are intensely active on social media, little is known about the possible underlying factors promoting the sharing of harmful or useful contents that could increase or decrease the likelihood of self-harm. Future investigations will concentrate on the following: (1)to measure behavioural attributes relating to social engagement (2) extensive analysis of self-harm-related emotion or sentiment (3) the language and linguistic styles used in posting self-harming content on Twitter (4) an analysis of the ego network of a self-harming user, and (5) to understand the frequency of social activity of the concerned users.